\documentclass[a4paper]{llncs}	
\usepackage{graphicx}
\usepackage{authblk}
\usepackage{todonotes}
\usepackage{setspace}
\usepackage{csquotes}
\usepackage{amssymb}
\usepackage{pifont}
\newcommand{\cmark}{\textcolor{green}{\ding{51}}}%
\newcommand{\xmark}{\textcolor{red}{\ding{55}}}%
\usepackage{url}
\usepackage{hyperref}

\let\figname=\figurename

\begin{document}

\title{Generation of Synthetic Images for Pedestrian Detection Using a Sequence of GANs}

  \author{Viktor Seib$^{1}$, Malte Roosen$^{1,2}$, Ida Germann$^{1,2}$, Stefan Wirtz$^{1}$, Dietrich Paulus$^{2}$}
  \institute{$^1$Motec GmbH, Oberweyerer Stra\ss e 21, 65589 Hadamar-Steinbach, Germany \\
   $^2$University of Koblenz, Universit\"atsstr. 1, 56070 Koblenz, Germany\\
  \email{\{viktor.seib,  stefan.wirtz\}@ametek.com \\ \{mroosen, idagermann, paulus\}@uni-koblenz.de}}

\maketitle

\begin{abstract}
Creating annotated datasets demands a substantial amount of manual effort. In this proof-of-concept work, we address this issue by proposing a novel image generation pipeline.
The pipeline consists of three distinct generative adversarial networks (previously published), combined in a novel way to augment a dataset for pedestrian detection. Despite the fact that the generated images are not always visually pleasant to the human eye, our detection benchmark reveals that the results substantially surpass the baseline.
The presented proof-of-concept work was done in 2020 and is now published as a technical report after a three years retention period.
\end{abstract}


\section{Introduction}

In the past decade, we could observe an enormous increase in performance in many computer visions tasks thanks to deep neural networks.
A substantial contribution to this achievement were large annotated datasets created by researchers worldwide.
However, the creation of annotated data demands a substantial amount of manual effort and therefore, large-scale datasets for supervised learning are only available for a limited range of applications \cite{deng2009imagenet}, \cite{lai2011large}, \cite{shahroudy2016ntu}, \cite{lin2014microsoft}, \cite{everingham2015pascal}.
In domains where such datasets are available, computers can perform vision tasks with (close to) human level accuracy \cite{russakovsky2015imagenet}.

Transfer learning allows to extend the application of trained networks into related vision domains.
To further enhance the performance in specific applications, additional data collection is often necessary.
While this is tedious but possible for many applications, other use cases depend on correct recognition of rare events with high confidence.
Large amounts of rare events, e.g.\ dangerous traffic situations or medical conditions, can not be collected that easily or under acceptable risks.

A current trend in neural network research therefore is to use synthetic data for training \cite{seib2020mixing}.
While real-world data is expensive to acquire and to annotate, synthetic data can be generated in arbitrary quantities once a suitable framework is properly set up.
Additionally, the annotations are generated along with the data itself.
Thus, synthetic data provides \enquote{perfect} ground truth information, since no manual and tedious -- and therefore error-prone -- annotation is required.
\\

For generating artificial and synthetic images that are almost indistinguishable from realistic photographs, generative adversarial networks (GANs) have proven to be particularly suitable \cite{curto2017high}.
However, in order to achieve good performance with training on synthetic data and avoid a domain shift problem, it is also necessary to properly reflect the feature distribution of the data with that of the target domain \cite{sankaranarayanan2018learning}. GANs address this issue by their capacity of image-to-image translation, allowing to transfer an image from one domain to another.

In the context of our work, transferring images from the domain of semantic maps to photo-realistic images \cite{wang2018high}, \cite{park2019semantic}, \cite{richter2022enhancing} is of special interest.
By additionally using the approach of Park et al.\ \cite{park2019semantic}, different styles for the resulting image can be chosen. This allows us style transfers and domain randomization and therefore can help to overcome the domain shift.

\begin{figure}[t!]
    \centering
    \includegraphics[width=\textwidth]{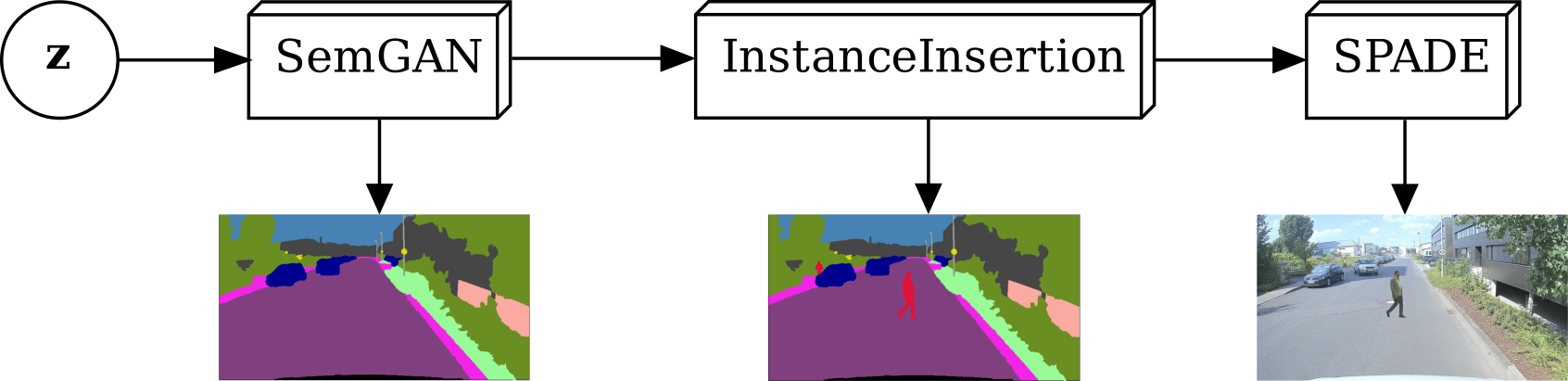}
    \caption{The augmentation pipeline consisting of three GANs that generate semantic maps and images from a latent variable $\mathbf{z}$. In the first step we use SemGAN \cite{ghelfi2019adversarial} to generate a semantic map. In the second step, the work proposed by Lee et al.\ \cite{lee2018context} is used to insert a new object instance (person) into the semantic map. Finally, SPADE \cite{park2019semantic} is used to convert the semantic map into an RGB image. Images used for illustration purposes, they are not the actual output.}
    \label{fig:AUP:pipeline}
\end{figure}

Images from the input domain of semantic maps highly resemble the ground truth annotation data of well-known urban datasets \cite{cordts2016cityscapes}, \cite{bdd}.
With the goal of building a dataset, we can view this as a dual problem: we have an annotation and need corresponding photo-realistic images. This idea is also used by Park et al.\ \cite{park2019semantic}.
However, the problem of obtaining the semantic annotation remains.
One could use existing ground truth annotations from different datasets, but this would constrain the scene layout to those present in the dataset. Another option is to synthesize the maps by another generative network, because semantic maps are also just images. Such an approach was e.g.\ introduced by Ghelfi et al.\ \cite{ghelfi2019adversarial}.

In this paper we combine these achievements of recent research and propose an augmentation pipeline for new types of synthetic datasets, illustrated in \figname \ref{fig:AUP:pipeline}.
We put our focus on the domain of urban traffic scenes and propose to synthesize semantic maps for this domain.
Our goal is to create a dataset for the purpose of pedestrian detection in urban settings. Therefore, the generated segmentation maps are augmented by inserting required object instances (here: pedestrians) and then, the resulting maps are translated into photo-realistic images.

Thus, the paper has three main contributions:
First, we study the generation of semantic maps using GANs.
Second, we propose a pipeline with three GANs which is capable of synthesizing semantic maps and converting them to photo-realistic training images.
Finally, we show that training an object detector benefits from additional data generated with this pipeline.
\\

In the following section, successful methods for data augmentation and common approaches in transfer learning are reviewed.
This includes data augmentation methods used in prevalent and successful neural networks. 
We also discuss different options typically used in transfer learning and fine-tuning and provide a short note on the feedforward design to calculate network weights.
Section\,\ref{pipeline} then gives a high-level overview of the developed pipeline.
Section\,\ref{impl} is the main part of this work. 
Therein we present technical details of the proposed pipeline and discuss the steps taken to sequentially combine different GANs.
In Section\,\ref{experiments} we describe the performed experiments and provide a discussion of the obtained the results.
Finally, Section\,\ref{summary} concludes this paper and gives an outlook to future work.


\section{Related Work}\label{related}

This section introduces common approaches to reuse trained networks available online for other purposes.
Further, we review methods to artificially increase the amount of data available for training without collecting or annotating additional images.

\subsection{Transfer Learning and Fine-Tuning}

A large variety of pretrained networks is available online in so called \enquote{model zoos}' for many frameworks and target devices. 
These networks are mostly pretrained on large datasets such as ImageNet \cite{deng2009imagenet}, Pascal VOC \cite{everingham2015pascal} or COCO \cite{lin2014microsoft}.
The size of these datasets allows the neural networks to learn abstract representations for many different object classes and embed generalized feature representations.

For the application of transfer learning and fine-tuning \cite{yosinski2014transferable}, \cite{oquab2014learning}, pretrained networks haven proven to be an effective way to achieve excellent classification and detection results despite limited data available for specific application domains.
Examples are traffic scene understanding in different lightning and weather conditions \cite{di2017cross},
anomaly detection in videos \cite{bansod2019transfer} and adapting traffic sign recognition from a large dataset to a specific region \cite{rosario2018deep}.
Transfer learning is also applicable to networks pretrained on generic images to medical image application, e.g.\ for skin cancer classification  \cite{esteva2017dermatologist}.
Surprisingly, transfer learning and fine-tuning can be applied successfully even when the data type and application domain of the target network significantly differ from the original pretrained network.
For instance, \cite{schwarz2015rgb} and \cite{eitel2015multimodal} use a network pretrained on RGB images to train a network on range image data.

\subsection{Synthetic Data}

Using synthetic data for training currently is very popular in neural network research. Scenarios specific to an application domain are typically modeled with 3D engines such as the Unreal Engine \cite{unrealengine} or Unity \cite{unitygameengine}.
The generated scenarios can be rendered as photo-realistic images from arbitrary perspectives and with arbitrary scene content and automatically generated ground truth annotations.

For urban traffic scenes, multiple synthetic datasets were proposed in the recent years.
A completely virtual city is presented in SYNTHIA \cite{ros2016synthia} as a photo-realistic image dataset, created with the Unity 3D engine.
As a virtual counterpart to KITTI \cite{geiger2013vision}, a popular real world dataset, VirtualKITTI \cite{gaidon2016virtual} was introduced.
Another approach is capturing photo-realistic images from video games \cite{drivingmatrix}, \cite{richter2017playing}. With video game approaches, ground truth data can also be generated automatically with a specific tool chain.
For a more detailed review on these datasets, please refer to \cite{seib2020mixing}.

\cite{ros2016synthia} and \cite{sadat2018effective} report improved results when a large amount of synthetic images is mixed with real images in training.
Specifically, for the task of semantic segmentation, the improvement occurs for foreground classes (pedestrians, cars, etc.) in contrast to background classes (sky, vegetation, etc.).
Both reports suggest that the improvements occur due to additional shape variations introduced with the synthetic data.
Typically, two main strategies are used to mix real and synthetic data in training:
While \cite{ros2016synthia} and \cite{richter2016playing} used mixed batches, \cite{drivingmatrix} and \cite{gaidon2016virtual} suggest pretraining a network with synthetic data and fine-tuning it with real data alone to better match the application domain.

More recently, a synthetic data generator for human-centric vision tasks has been proposed in \cite{ebadi2021peoplesanspeople}.
The authors also follow the strategy of pretraining with synthetic data and fine-tuning with domain specific real data.

A common concern when training a network on synthetic images is the degrading performance on real images.
This performance gap is referred to as domain shift and is linked to synthetic data having less variation in appearance as real data \cite{drivingmatrix}, \cite{sadat2018effective}.
One option to address this issue is domain randomization.
In \cite{tremblay2018training}, the authors propose to create synthetic images that do not look photo-realistic at all.
They generate cars with random, unrealistic parameters for lighting, pose and textures.
The idea thereby is that the network has to learn to detect objects independently from their texture.

Contrarily, it is also possible to use photo-realistic models and textures, e.g.\ as synthetic data generator for humans \cite{ebadi2021peoplesanspeople}.
The randomization is applied to the environmental properties such as background, lighting, occlusions and random objects.
This allows a network to learn to detect people independently from any context.
Further, the realistic textures help avoiding false positive detections based on shapes only, such as shadows.

An advanced option to address the domain shift is applying GANs for image-to-image translation, further described in the following subsection.

\subsection{Generative Adversarial Networks}

Generative Adversarial Networks (GANs) \cite{goodfellow2014generative}, \cite{radford2015unsupervised} have gained recent attention in many fields, for example in the generation of photo-realistic face images \cite{karras2017progressive}, \cite{curto2017high}
3D shapes \cite{wu2016learning} or style transfer \cite{johnson2016perceptual}.
In the same manner as a GAN can be trained to generate faces, it can also be trained to generate arbitrary images to extend a dataset.
This was also demonstrated in the domain of medical images, typically regarded as a domain with only few available data instances \cite{frid2018gan}.

There have also been attempts to create complex semantic maps with GANs \cite{ghelfi2019adversarial}.
However, while images are in a rather continuous color space, semantic maps only contain the discrete class labels.
As GANs are known to be difficult to train on discrete distributions, generating semantic maps with them is a challenging task.
Further, the work presented in \cite{lee2018context} generates realistic 2D silhouettes of pedestrians and is capable of placing them inside semantic maps.
While these examples open many possibilities, training a GAN is challenging and requires large amounts of images.
If the available dataset is small it will not suffice to train a GAN to generate new images.

When using synthetic images to extend a dataset, style transfer can be used for domain randomization, addressing the previously described domain shift problem.
GANs for style transfer replace the style of one image with that of another, all while preserving the original image content \cite{johnson2016perceptual}, \cite{gatys2016}.

GANs can not only transfer a style, but they are also capable of transferring images directly from one domain to another \cite{zhu2017unpaired}.
For autonomous driving, images can be transferred between day and night, summer and winter, and sunny and rainy weather \cite{shorten2019survey}.
This method is not limited to synthetic data and can also be applied to real images, e.g.\ to train a car to drive at rainy weather when the dataset contains sunny images.
Domain transfer does not need paired image examples, but still requires a sufficiently large dataset to train domains of interest.

Of special interest in this work is domain transfer from semantic maps to photo-realistic images \cite{park2019semantic}, \cite{wang2018high}, \cite{richter2022enhancing}.
The idea is to use a GAN for generating semantic maps and then transferring these maps to photo-realistic images.
The details of our pipeline are outlined in the following section.

\section{Pipeline Overview}\label{pipeline}

Our proposed pipeline (see \figname\,\ref{fig:AUP:pipeline}) consists of three steps and every step is represented by a distinct GAN.
We make careful adaptations to be able to use the output of one step as the input of the subsequent pipeline step.
Table\,\ref{tab:overview} gives an overview of the different GANs involved in the pipeline and possible output images.

\begin{table}
\caption{Pipeline overview with possible output images per step. The images are used for illustration purposes and do not represent the actual output generated by the pipeline.}
\label{tab:overview}
\begin{center}
\begin{tabular}{c||c|c|c}
Pipeline step     & Step 1 & Step 2 & Step 3\\
\hline
\hline
Relies on work by & Ghelfi et al.\ \cite{ghelfi2019adversarial} & Lee et al.\ \cite{lee2018context} & Park et al.\ \cite{park2019semantic}\\
Code online       & \xmark & \cmark & \cmark \\
Weights online    & \xmark & \xmark & \cmark \\
Possible output   & \includegraphics[width=0.25\textwidth]{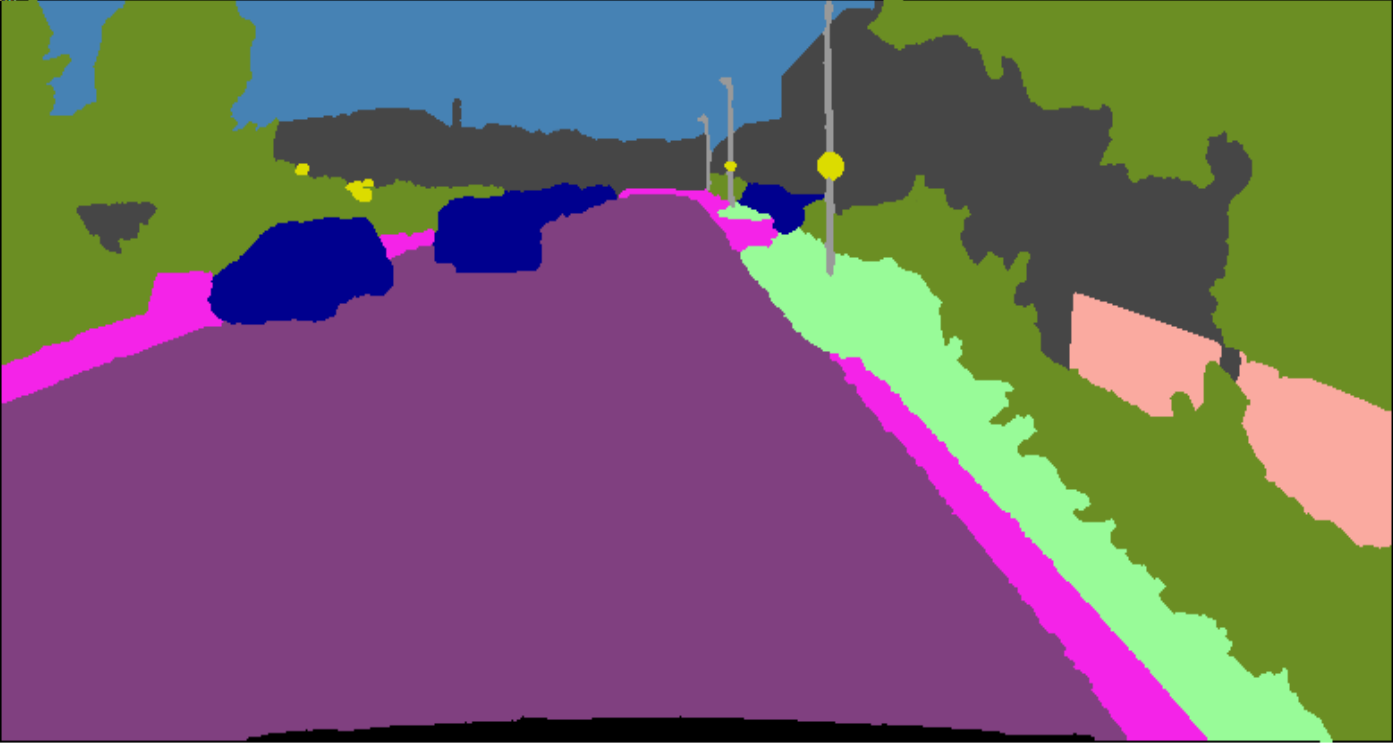} & \includegraphics[width=0.25\textwidth]{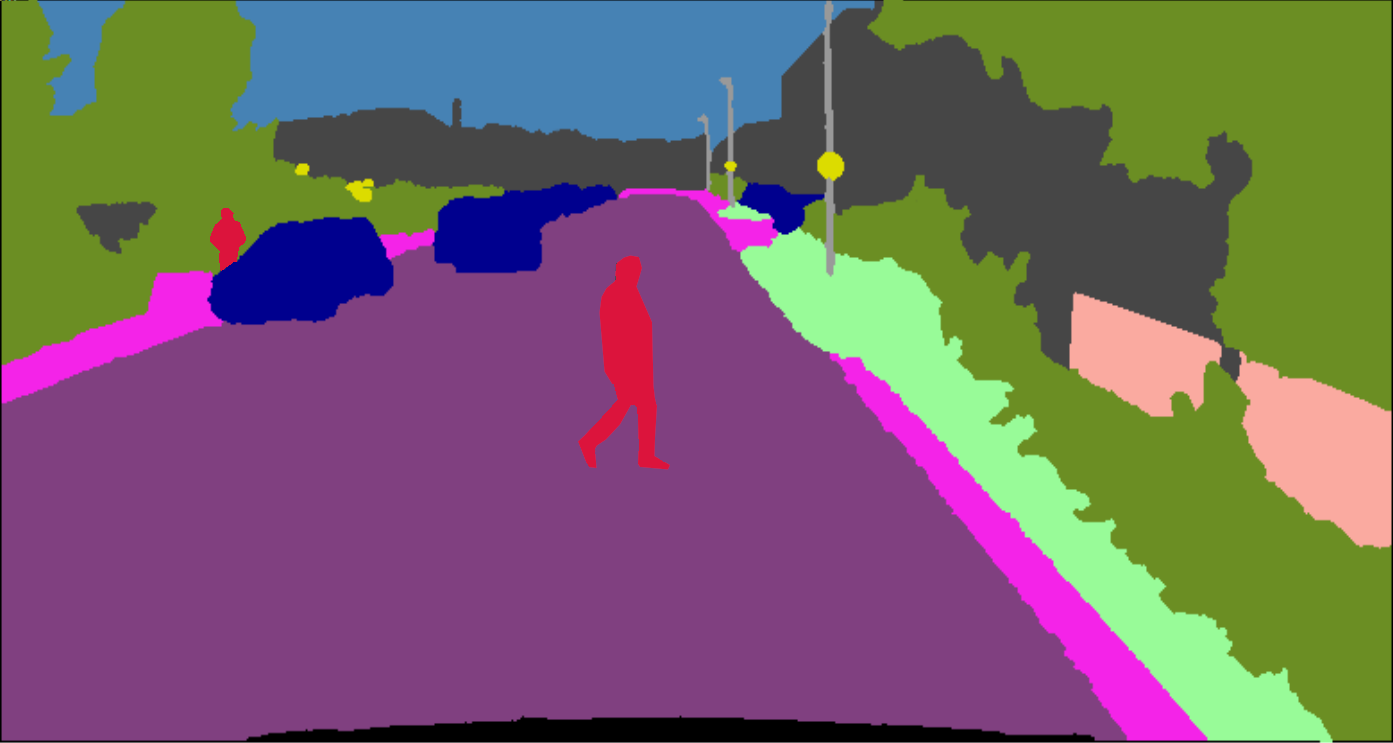} & \includegraphics[width=0.25\textwidth]{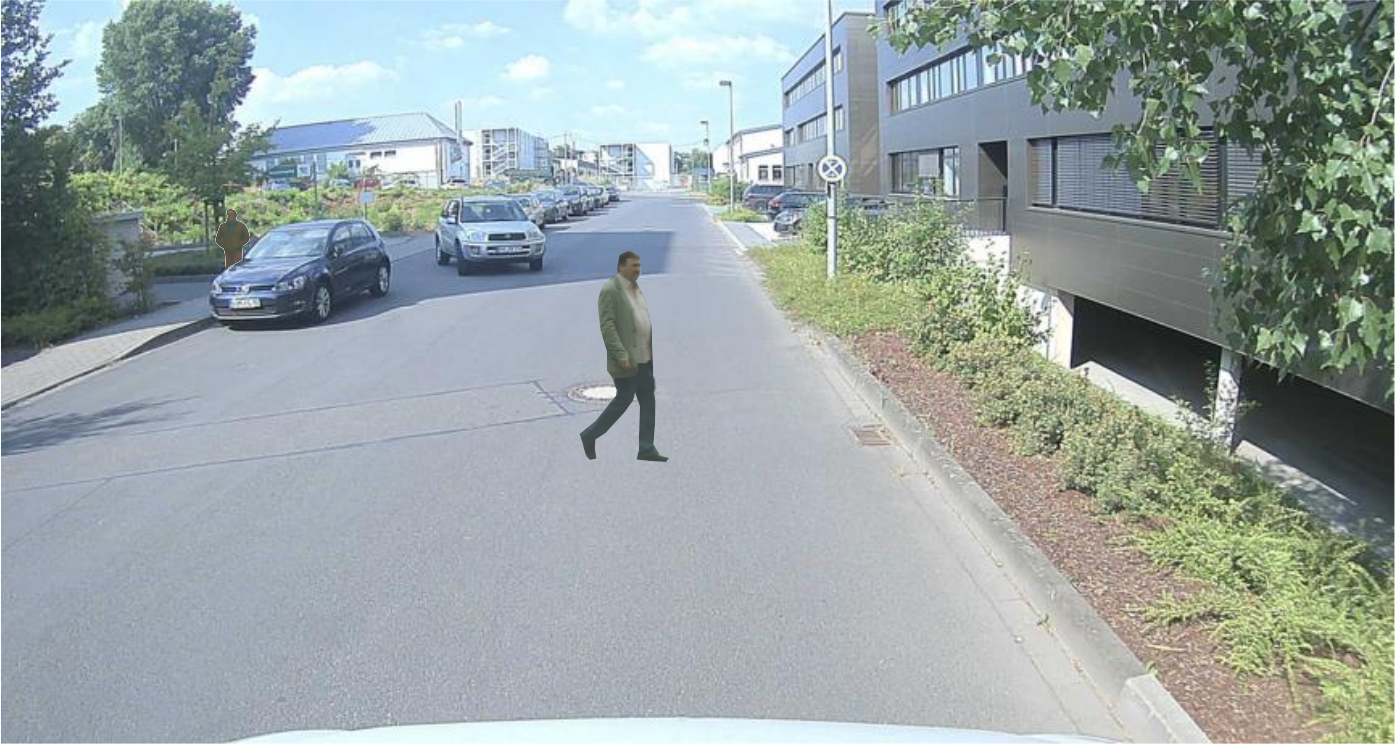}
\end{tabular}
\end{center}
\end{table}

The first step of our pipeline is used to generate semantic maps of urban traffic scenes.
We build upon the work of Ghelfi et al.\ \cite{ghelfi2019adversarial}.
However, since neither their code nor their trained network weights are available online, we implement their approach with several adaptations as described in Section\,\ref{impl}.
The second step relies on the work of Lee et al.\ \cite{lee2018context} and takes the semantic maps from the first step as input to generate realistic instances at appropriate locations.
These are inserted into the semantic maps and are used to generate more objects of interest for the target task (here: pedestrians).
We make only small modifications to the code available online and train the GAN from scratch.
Finally, with the code and weights available for the third step, we use the GAN as provided by Park et al.\ \cite{park2019semantic} to transfer the semantic maps with inserted object instances to photo-realistic images.
Thus, we obtain photo-realistic images with corresponding semantic maps that we can use to extend an available dataset for pedestrian detection.

Since no trained networks weights are provided, we have to train the first two pipeline steps from scratch.
For this purpose, we use the Cityscapes dataset \cite{cordts2016cityscapes}.
We use 5000 images for training and all 34 classes available in Cityscapes.

\section{Pipeline Details}\label{impl}

This section describes the different pipeline steps in detail and outlines the adaptations applied to the images for sequential processing.

\subsection{Step 1: Generating Semantic Maps}
\label{step_semgan}

The first GAN in the pipeline is based on the work of Ghelfi et al.\ \cite{ghelfi2019adversarial}, dubbed SemGAN, and is used to generate semantic maps.
The difficulty of synthesizing semantic maps arises from their discrete nature.
Pixels in semantic maps can take only one of the class ids as their values.
While a slight change in value will not significantly alter the meaning of a pixel in a color image, it will change the class of that pixel in a semantic map.
Further, representing classes with consecutive numbers implies an ordering between them which usually does not exist.
We therefore have to use one-hot encoding to represent the pixel data.
Each pixel is expanded to a $k$-dimensional vector with values in the range $[0, 1]$, where  $k$ is the number of classes.
A pixel belonging to the class with id $i$ is represented as a vector of zeros with only the $i$-th element set to one.
The encoded semantic map has the shape $(w, h, k)$, where $w, h$ are the width and height of the semantic map.
The last channel can also be interpreted as a probability distribution for every pixel over the classes.
Ghelfi et al.\ use this encoding in their GAN to represent semantic maps.
The generator outputs data of the aforementioned shape and consequently the discriminator's input has this shape as well.
To normalize the generator's output, they use softmax instead of the usual tanh as the final activation layer.
This approach allows Ghelfi et al.\ to generate semantic maps of up to 128 $\times$ 128 pixels.

\begin{figure}
 \centering
 \includegraphics[width=\textwidth]{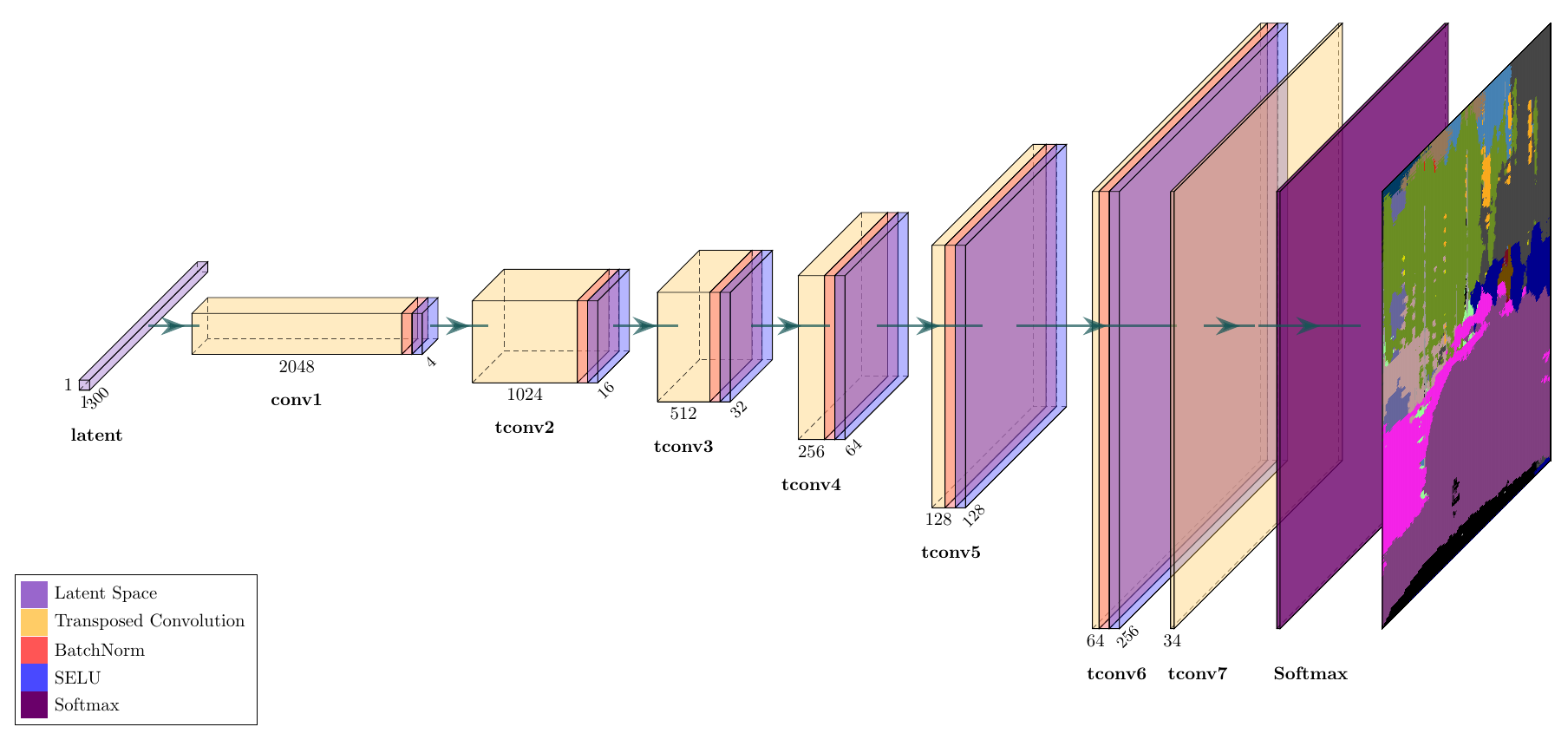}
 \caption{Our implementation of the SemGAN generator with the described modifications.}
 \label{semgan-generator}
\end{figure}

\begin{figure}
 \centering
 \includegraphics[width=\textwidth]{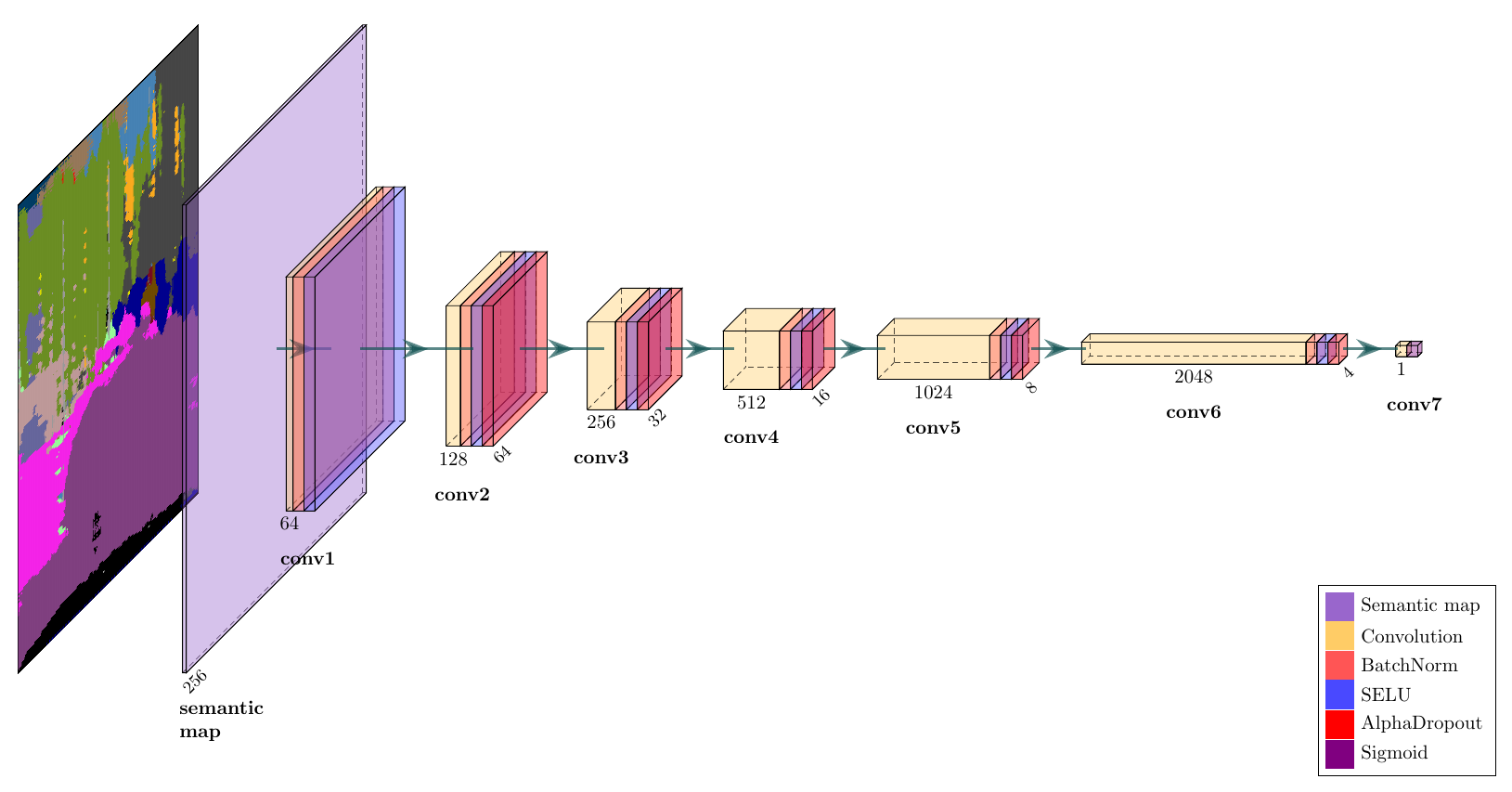}
 \caption{Our implementation of the SemGAN discriminator with the described modifications.}
 \label{semgan-discriminator}
\end{figure}

We adapt some modifications to the approach described by Ghelfi et al.\ as described below.
The resulting generator and discriminator are shown in \figname\,\ref{semgan-generator} and in \figname\,\ref{semgan-discriminator}.
We follow the guidelines presented in \cite{radford2015unsupervised} to improve training stability and performance of convolutional GANs.
Our modification include replacing all remaining linear layers by convolutional layers and adding a dropout layer after each block in the discriminator.

Since the GANs in subsequent steps of our pipeline use higher resolutions, we create bigger semantic maps.
To increase the output resolution of SemGAN we modify its architecture as suggested by Curto et al.\ in \cite{curto2017high}.
This includes replacing (leaky) ReLUs by scaled exponential linear units (SELUs) as activation layers.
Curto et al.\ observe that SELUs in combination with batch normalization improve convergence speed as well as training stability.
Further, as suggested by Klambauer et al.\ \cite{klambauer2017self}, we replace dropout layers by AlphaDropouts.
These have the property of preserving the self-normalization of SELUs.

We train the SemGAN generator with feature matching loss \cite{salimans2016improved} and the discriminator with binary cross entropy loss.
Feature matching loss computes the error on internal feature layers instead of the final output of the discriminator.
We use the output from the 6th convolution (see ''conv6'' in \figname\,\ref{semgan-discriminator}).
During an ablation study we confirmed that using binary cross entropy for both networks makes training unstable.

\begin{figure}[t!]
 \centering
 \includegraphics[width=0.31\textwidth]{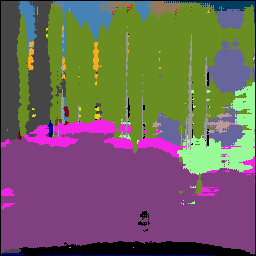}
 \includegraphics[width=0.31\textwidth]{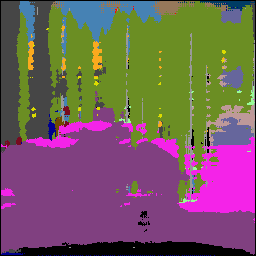}
 \includegraphics[width=0.31\textwidth]{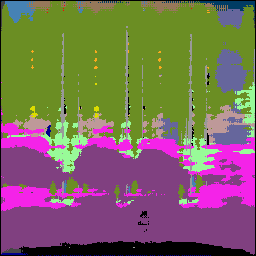}
 \caption{Semantic maps generated by the SemGAN network (output of the first pipeline step).}
 \label{semgan-images}
\end{figure}

With our modifications, the SemGAN generator outputs a one-hot encoded tensor of shape $(256 \times 256 \times 34)$ which can be decoded into a semantic map by applying $\mathrm{argmax}$ over the last channel.
Example semantic maps generated in this pipeline step are shown in \figname\,\ref{semgan-images}.
The seemingly poor image quality stems from the few training images used in this step.
Training SemGAN for 500 epochs took 18 hours on two NVIDIA GTX 1080 Ti GPUs with a batch size of 32.
We use two Adam optimizers, one for the generator and one for the discriminator.
Both are set to the same parameters: A learning rate of $0.0002$, $\beta_1 = 0.3$ and $\beta_2 = 0.999$.
Further, we use a dropout rate of $0.3$ for the AlphaDropouts.

\subsection{Step 2: Inserting Instances}
\label{step_insins}
The generated semantic maps largely consist of background object classes like roads, buildings and vegetation.
On the other hand, foreground objects like people or bicycles are rarely generated.
We therefore add the instance insertion step to ensure that sufficient pedestrians are present in the generated data.

\begin{figure}[b!]
 \centering
 \includegraphics[width=0.6\textwidth]{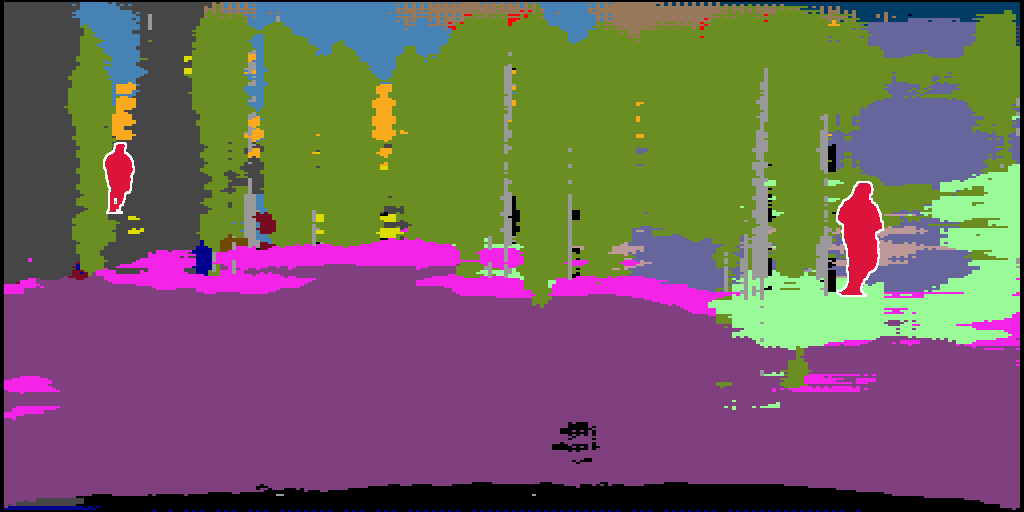}\\
 \includegraphics[width=0.6\textwidth]{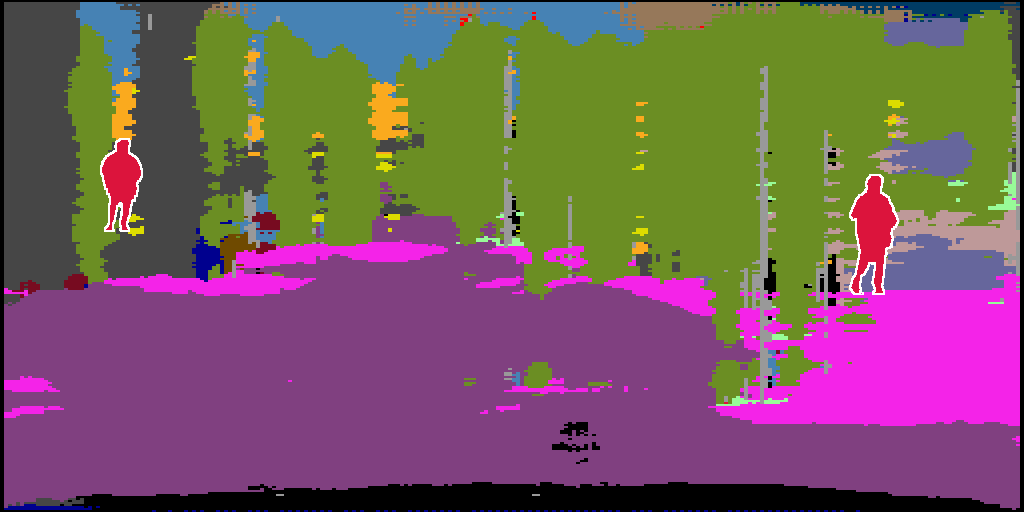}\\
 \includegraphics[width=0.6\textwidth]{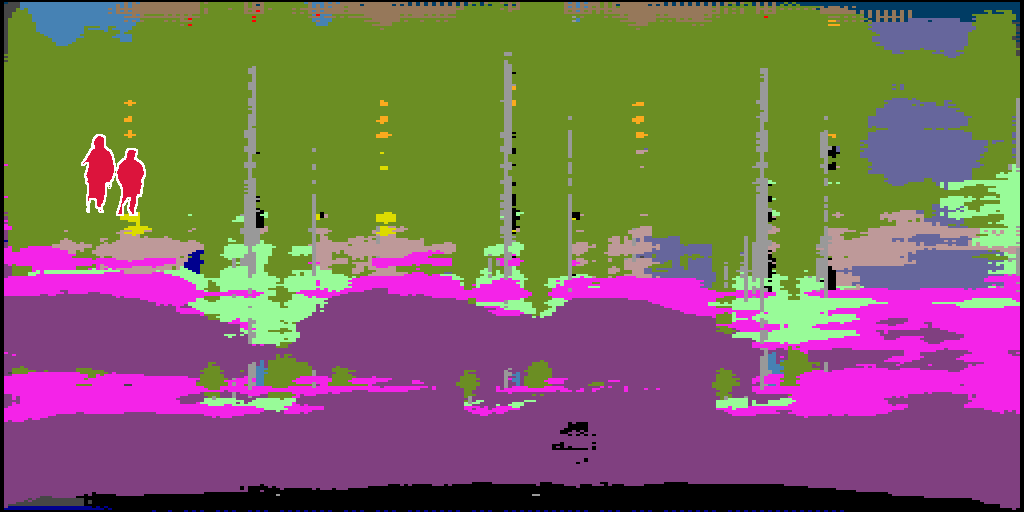}
 \caption{Semantic maps with inserted person instances (output of the second pipeline step). Inserted instances are highlighted by a white border for visualization purposes.}
 \label{instance-images}
\end{figure}

The approach of Lee et al.\ \cite{lee2018context} uniquely fits this task.
It consists of the \enquote{what} and \enquote{where} modules and generates silhouettes of instances belonging to trained classes at realistic positions in semantic maps.
We integrated the implementation \cite{instancegithub} published alongside the paper into our system and trained it for the class person using the default parameters for 100 epochs.
Training time was around two days on a single NVIDIA GTX 1080 Ti at a batch size of one.

The instance insertion step takes a semantic map of size $(1024 \times 512)$ as input.
We resize the output semantic maps from step 1 of the pipeline without any interpolation in order not to produce invalid class labels in the output.
The \enquote{where} module works on a version downsampled to $(256 \times 128)$.
The full scale version map is used by the \enquote{what} module which outputs the shape of the new instance as a $(128 \times 128)$ binary mask.
Example semantic maps from the first step, resized and with added pedestrian instances are shown in \figname\,\ref{instance-images}.

\subsection{Step 3: SPADE}
\label{step_spade}

\begin{figure}[b!]
 \centering
 \includegraphics[width=0.6\textwidth]{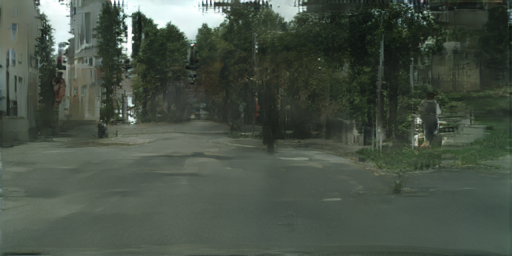}
 \includegraphics[width=0.6\textwidth]{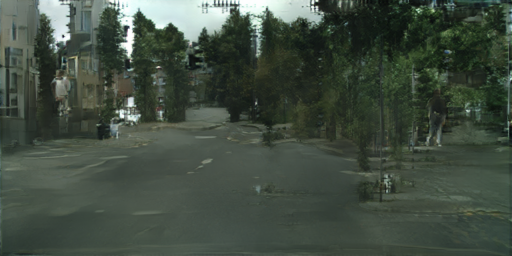}
 \includegraphics[width=0.6\textwidth]{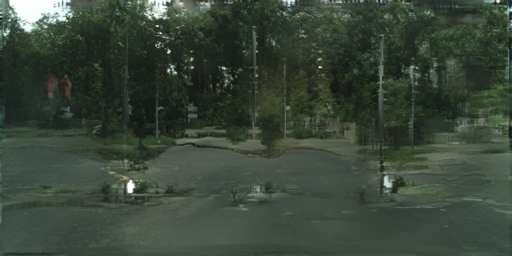}
 \caption{Generated semantic maps converted to photo-realistic images (output of the third pipeline step).}
 \label{spade-images}
\end{figure}

The third and final step of our pipeline is formed by the SPADE generator.
It allows us to turn the previously generated semantic maps into the photo-realistic color images needed to train the object detection network.
Park et al.\ \cite{park2019semantic} published their code along with pretrained weights \cite{spadegithub}.
The network was pretrained for 200 epochs with a batch size of 32.
We integrated this implementation into our pipeline and used the provided weights.

Park et al.\ combine the generator with a multi-scale discriminator. 
Multi-scale means that there are multiple discriminators operating with differently scaled versions of the input.
In the case of SPADE and the Cityscapes dataset, there are two discriminators: one for the full-sized input image and one downsampled to half the size.
SPADE works with semantic maps and images of the size $(512 \times 256)$.
Some example output images from our pipeline are shown in \figname\,\ref{spade-images}.

\section{Experiments and Discussion}\label{experiments}

The goal of this work is to generate synthetic training data to augment a relatively small real world training dataset.
We are interested in detecting pedestrians in urban settings.
For our experiments, we use the popular YOLOv3 detection network \cite{yolov3}.
We leave the backbone unchanged and retrain the detector part of the network with 100,000 synthetic images acquired from an external service provider. 
Subsequently, we fine-tune the detector with different datasets to examine the effects of the synthetic data generated with our approach.
Table\,\ref{tab:train_data} provides an overview of the datasets used for fine-tuning.

\begin{table}
\caption{The research datasets encompass the Cityscapes dataset (5000 images) \cite{cordts2016cityscapes} and the BDD dataset \cite{bdd}.}
\label{tab:train_data}
\begin{center}
\begin{tabular}{l||c}
Dataset & \# training images \\
\hline
Motec Data        & 10,500 \\
Research Data & 105,000  \\
GAN Sequence (this work) & 5,000 \\

\end{tabular}
\end{center}
\end{table}

\begin{figure}[t!]
 \centering
 \includegraphics[width=0.4\textwidth]{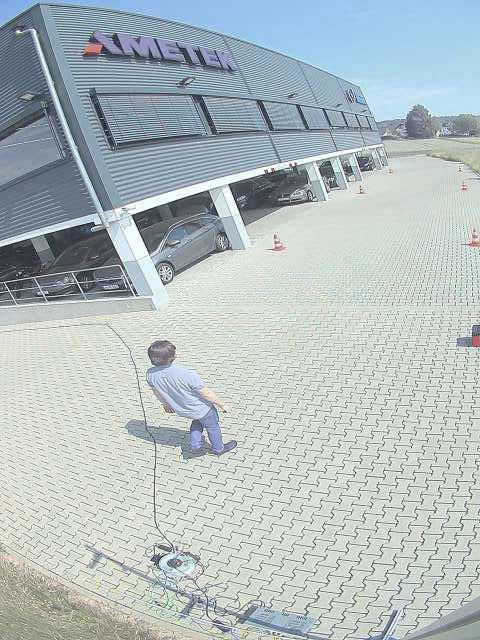}
 \includegraphics[width=0.4\textwidth]{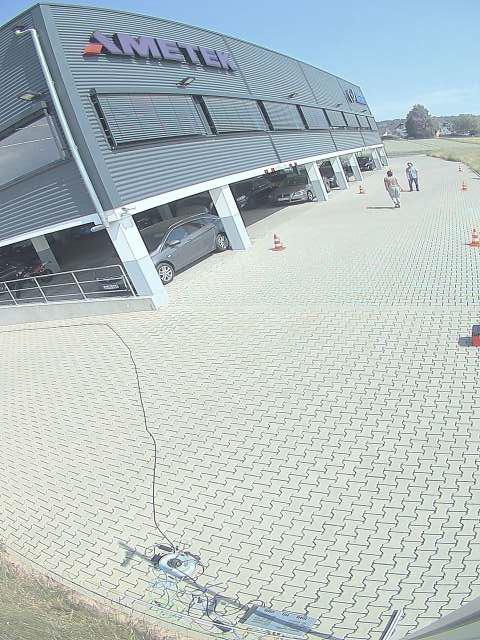}
 \caption{Exemplary training images from one of the sequences with a near range image (left) and a far range image (right).}
 \label{fig:example_images}
\end{figure}

The resulting networks are tested on a proprietary dataset captured by Motec.
These datasets contain image sequences containing one or multiple people.
The people move freely and in different distances from the camera.
Each sequence is divided into near range and far range images, depending on the position of the people.
We define near range as everything closer than 10\,m to the camera, while far range encompasses a distance of 10\,m to 20\,m.
Example images are shown in \figname\,\ref{fig:example_images}.

Table\,\ref{tab:results} shows the detection results for the different fine-tuned networks.
To compute the metrics, we count a true positive detection if a bounding box of a detection hypothesis has an overlap of at least 50\% with an annotated box.
Comparing the first two lines of Table\,\ref{tab:results}, we observe that adding the generated images greatly improves over the training results with real data only.
The F-Score increases by almost 10 pp for near range images and by about 12 pp for far range images.
More importantly, the recall greatly improves in both scenarios, indicating that less pedestrians are overseen by the resulting network.
On the other hand, when using only the generated images, the results on near range images fall below the results when training with real data only.
However, for far range images we still observe a small improvement.
This indicates that the real world training data is rather centered on near range images, while the generated images also encompass many far range images.

As an ablation study we also fine-tune the detector with research datasets alone and including the generated images (see last two lines of Table\,\ref{tab:results}).
The results obtained with research datasets are worse than with Motec data, despite having much more training images.
This is because the Motec data is specifically tailored to our use-case, e.g.\ in terms of composition of scenes, camera height and inclination.
Again, we observe better results when our generated images are added to the training data.
Further, the highest overall precision on far range images is obtained when using research datasets and generated images together.

In an additional ablation study we performed the same experiments without retraining the YOLOv3 detection layers with the 100,000 synthetic images from our external service provider.
This means that we directly fine-tuned the detection layers of YOLOv3 with the indicated datasets.
The overall results were significantly worse than with the additional retraining.
Further, while adding the generated images to the train set improved the precision, the recall results dropped resulting in an overall \textit{lower} f-score than in the first case.

\begin{table}
\caption{Detection results obtained after retraining the YOLOv3 detector with different datasets. The "GAN Sequence" data is generated by the proposed pipeline in this work.}
\label{tab:results}
\begin{center}
\begin{tabular}{l||ccc|ccc}
Detector & \multicolumn{3}{c|}{Near Range Images}& \multicolumn{3}{c}{Far Range Images} \\
Finetuned With     & F-Score & Precision & Recall & F-Score & Precision & Recall\\
\hline
Motec Data                           & 76.9 & 90.0 & 67.2 & 61.8 & 88.0 & 47.6 \\
Motec Data + GAN Seq.        & \textbf{86.1} & \textbf{98.0} & \textbf{76.7} & \textbf{73.9} & 90.2 & \textbf{62.6} \\
GAN Sequence                    & 52.6 & 66.2 & 43.6 & 62.6 & 78.7 & 51.9 \\
\hline                               
Research Data                    & 26.5 & 36.9 & 20.7 & 49.4 & 93.4 & 33.6 \\
Research Data + GAN Seq. & 32.3 & 69.6 & 21.0 & 54.9 & \textbf{94.7} & 38.7 \\
\end{tabular}
\end{center}
\end{table}

When inspecting the images generated by our pipeline, we need to admit that they are not visually pleasant.
Further, most images look alike and exhibit artifacts from different classes.
Since the first pipeline step defines the overall image layout, it is the weak point of the whole pipeline.
Obviously, it needs much more training data and some more architectural adjustments.

\begin{figure}
 \centering
 \includegraphics[width=0.48\textwidth]{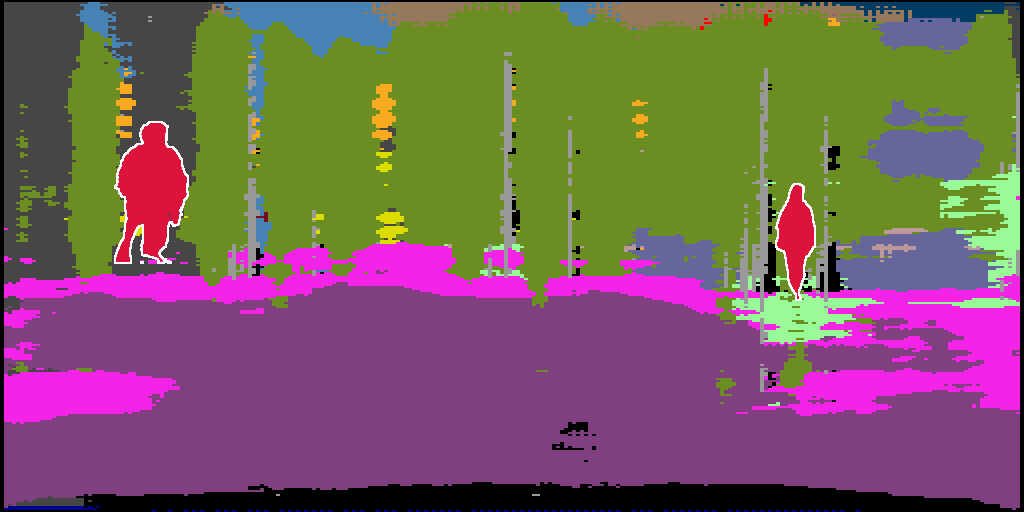}
 \includegraphics[width=0.48\textwidth]{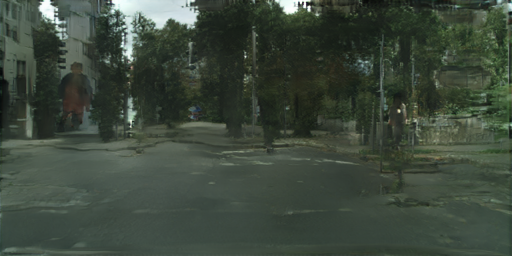}
 \includegraphics[width=0.48\textwidth]{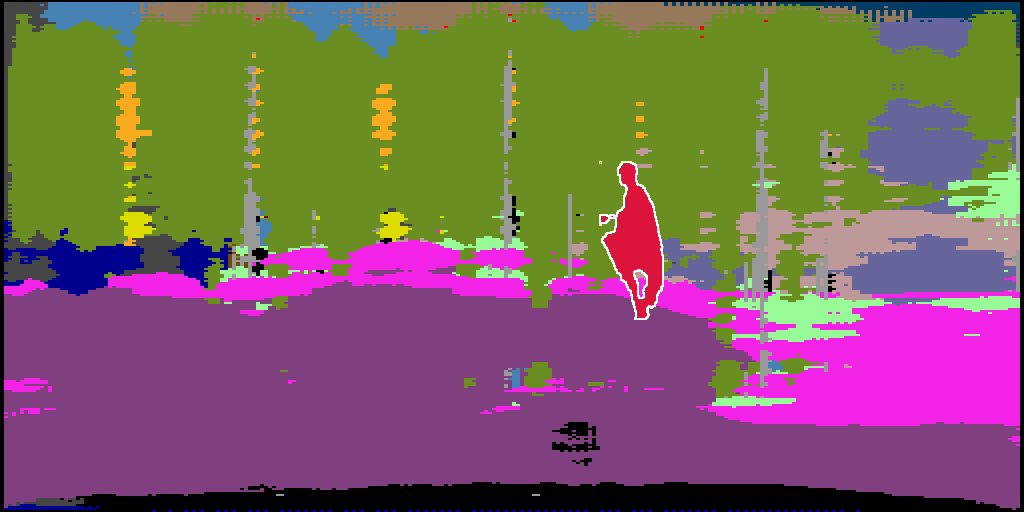}
 \includegraphics[width=0.48\textwidth]{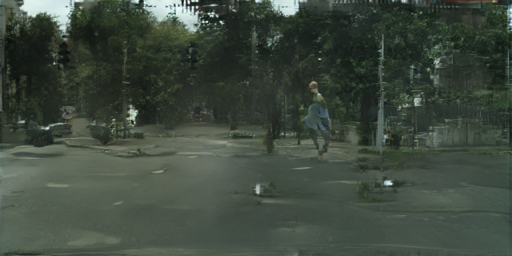}
 \includegraphics[width=0.48\textwidth]{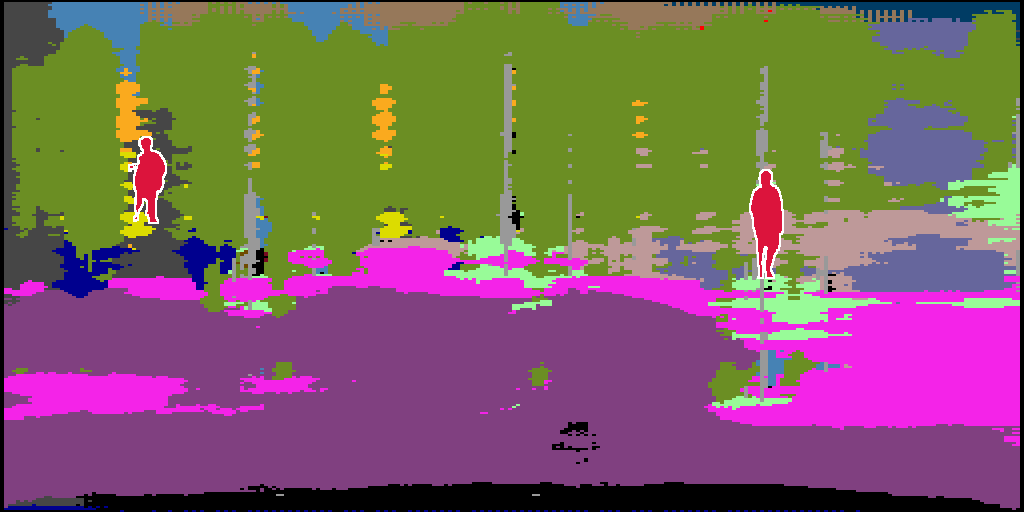}
 \includegraphics[width=0.48\textwidth]{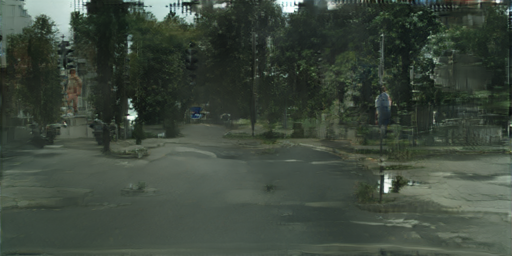}
 \includegraphics[width=0.48\textwidth]{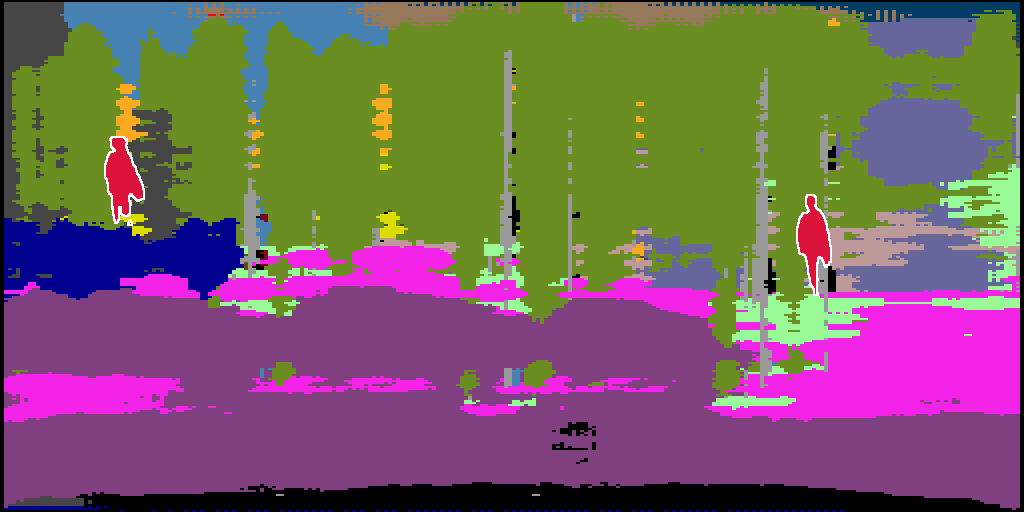}
 \includegraphics[width=0.48\textwidth]{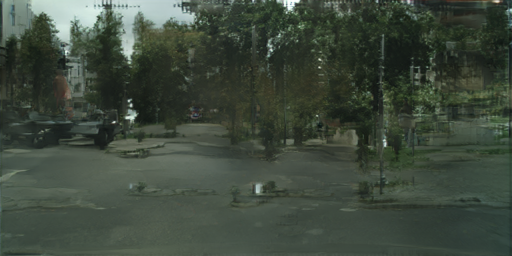}
 \includegraphics[width=0.48\textwidth]{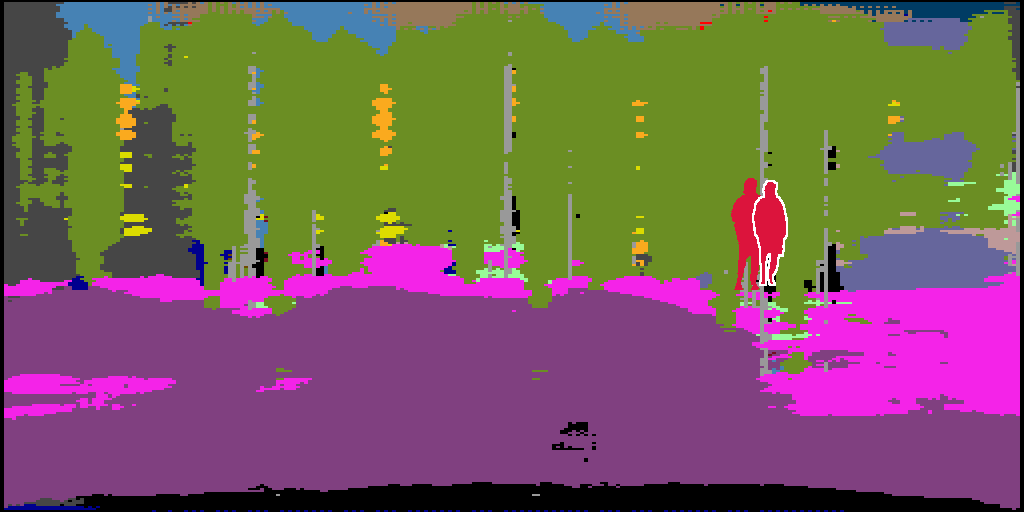}
 \includegraphics[width=0.48\textwidth]{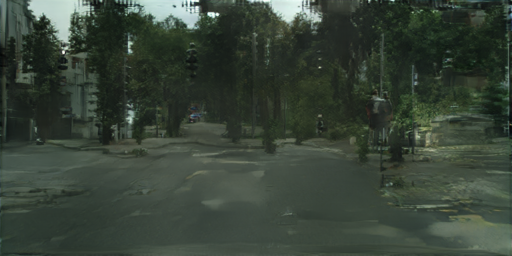}
 \includegraphics[width=0.48\textwidth]{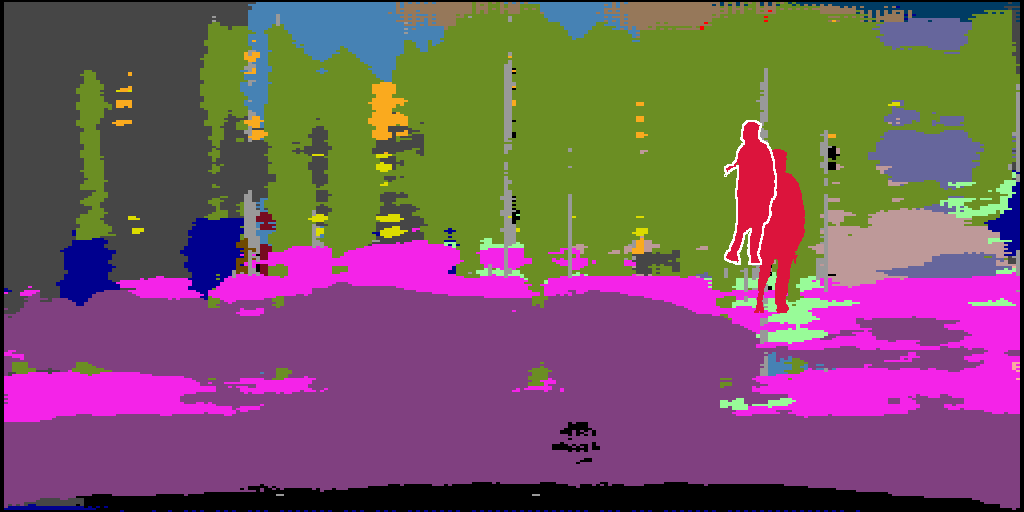}
 \includegraphics[width=0.48\textwidth]{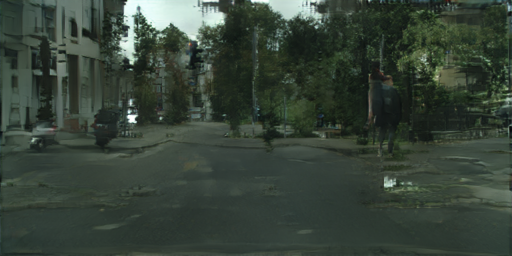}
 \includegraphics[width=0.48\textwidth]{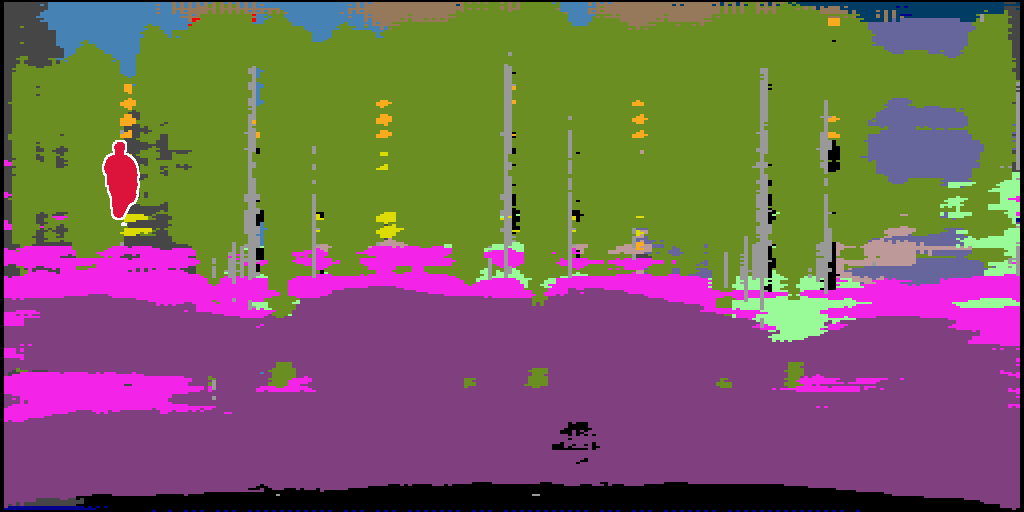}
 \includegraphics[width=0.48\textwidth]{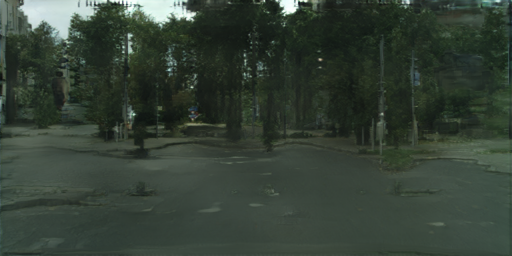}
 \caption{More example images: semantic maps with inserted instances (left column) and corresponding photo-realistic images (right column).}
 \label{more-images}
\end{figure}

Nevertheless and despite the poor image quality, we could obtain a substantial improvement in the detection evaluation.
This is likely because the GANs emphasize important object features which are not necessarily visually pleasant to the human eye.
Additional images generated by our pipeline are shown in \figname\,\ref{more-images}.

\section{Summary}\label{summary}

In this work, a pipeline for synthetic image generation that consists of three distinct generative adversarial networks (GANs) is introduced.
Each of the steps employs a GAN. The steps are sequentially combined to enhance a dataset for pedestrian detection through the generation of novel images.
The first two pipeline steps are used to generate a semantic map and to add additional objects instances of interest. Then, the semantic maps are converted into photo-realistic images.

In future work we plan to improve the architecture of the first pipeline step which we believe to be the weak point of the current pipeline.
Further, we plan to collect data to train the first two pipeline steps and no longer depend on research datasets for the generation of synthetic images.
Despite these shortcomings the results obtained in this proof of concept study are encouraging.
By using the generated data we could substantially improve the detection results in our target application domain.

\bibliographystyle{splncs03.bst}
\bibliography{references}

\begin{thebibliography}{10}
\providecommand{\url}[1]{\texttt{#1}}
\providecommand{\urlprefix}{URL }

\bibitem{bansod2019transfer}
Bansod, S., Nandedkar, A.: Transfer learning for video anomaly detection.
  Journal of Intelligent \& Fuzzy Systems  36(3),  1967--1975 (2019)

\bibitem{cordts2016cityscapes}
Cordts, M., Omran, M., Ramos, S., Rehfeld, T., Enzweiler, M., Benenson, R.,
  Franke, U., Roth, S., Schiele, B.: The cityscapes dataset for semantic urban
  scene understanding. In: Proceedings of the IEEE conference on computer
  vision and pattern recognition. pp. 3213--3223 (2016)

\bibitem{curto2017high}
Curt{\'o}, J.D., Zarza, I.C., De~La~Torre, F., King, I., Lyu, M.R.:
  High-resolution deep convolutional generative adversarial networks. arXiv
  preprint arXiv:1711.06491  (2017)

\bibitem{deng2009imagenet}
Deng, J., Dong, W., Socher, R., Li, L.J., Li, K., Fei-Fei, L.: Imagenet: A
  large-scale hierarchical image database. In: 2009 IEEE conference on computer
  vision and pattern recognition. pp. 248--255. Ieee (2009)

\bibitem{di2017cross}
Di, S., Zhang, H., Li, C.G., Mei, X., Prokhorov, D., Ling, H.: Cross-domain
  traffic scene understanding: A dense correspondence-based transfer learning
  approach. IEEE transactions on intelligent transportation systems  19(3),
  745--757 (2017)

\bibitem{ebadi2021peoplesanspeople}
Ebadi, S.E., Jhang, Y.C., Zook, A., Dhakad, S., Crespi, A., Parisi, P.,
  Borkman, S., Hogins, J., Ganguly, S.: Peoplesanspeople: A synthetic data
  generator for human-centric computer vision  (2021)

\bibitem{eitel2015multimodal}
Eitel, A., Springenberg, J.T., Spinello, L., Riedmiller, M., Burgard, W.:
  Multimodal deep learning for robust rgb-d object recognition. In: 2015
  IEEE/RSJ International Conference on Intelligent Robots and Systems (IROS).
  pp. 681--687. IEEE (2015)

\bibitem{unrealengine}
{Epic Games}: Unreal engine, \url{https://www.unrealengine.com}

\bibitem{esteva2017dermatologist}
Esteva, A., Kuprel, B., Novoa, R.A., Ko, J., Swetter, S.M., Blau, H.M., Thrun,
  S.: Dermatologist-level classification of skin cancer with deep neural
  networks. Nature  542(7639),  115--118 (2017)

\bibitem{everingham2015pascal}
Everingham, M., Eslami, S.A., Van~Gool, L., Williams, C.K., Winn, J.,
  Zisserman, A.: The pascal visual object classes challenge: A retrospective.
  International journal of computer vision  111(1),  98--136 (2015)

\bibitem{frid2018gan}
Frid-Adar, M., Klang, E., Amitai, M., Goldberger, J., Greenspan, H.: Gan-based
  data augmentation for improved liver lesion classification  (2018)

\bibitem{gaidon2016virtual}
Gaidon, A., Wang, Q., Cabon, Y., Vig, E.: Virtual worlds as proxy for
  multi-object tracking analysis. In: Proceedings of the IEEE conference on
  computer vision and pattern recognition. pp. 4340--4349 (2016)

\bibitem{gatys2016}
Gatys, L.A., Ecker, A.S., Bethge, M.: Image {{Style Transfer Using
  Convolutional Neural Networks}}. In: 2016 {{IEEE Conference}} on {{Computer
  Vision}} and {{Pattern Recognition}}. pp. 2414--2423. {IEEE Computer
  Society}, {Las Vegas, NV, USA} (Jun 2016)

\bibitem{geiger2013vision}
Geiger, A., Lenz, P., Stiller, C., Urtasun, R.: Vision meets robotics: The
  kitti dataset. The International Journal of Robotics Research  32(11),
  1231--1237 (2013)

\bibitem{ghelfi2019adversarial}
Ghelfi, E., Galeone, P., De~Simoni, M., Di~Mattia, F.: Adversarial pixel-level
  generation of semantic images. arXiv preprint arXiv:1906.12195  (2019)

\bibitem{goodfellow2014generative}
Goodfellow, I., Pouget-Abadie, J., Mirza, M., Xu, B., Warde-Farley, D., Ozair,
  S., Courville, A., Bengio, Y.: Generative adversarial nets. In: Advances in
  neural information processing systems. pp. 2672--2680 (2014)

\bibitem{johnson2016perceptual}
Johnson, J., Alahi, A., Fei-Fei, L.: Perceptual losses for real-time style
  transfer and super-resolution. In: European conference on computer vision.
  pp. 694--711. Springer (2016)

\bibitem{drivingmatrix}
Johnson-Roberson, M., Barto, C., Mehta, R., Sridhar, S.N., Rosaen, K.,
  Vasudevan, R.: Driving in the matrix: Can virtual worlds replace
  human-generated annotations for real world tasks? arXiv preprint
  arXiv:1610.01983  (2016)

\bibitem{karras2017progressive}
Karras, T., Aila, T., Laine, S., Lehtinen, J.: Progressive growing of gans for
  improved quality, stability, and variation. arXiv preprint arXiv:1710.10196
  (2017)

\bibitem{klambauer2017self}
Klambauer, G., Unterthiner, T., Mayr, A., Hochreiter, S.: Self-normalizing
  neural networks. Advances in neural information processing systems  30 (2017)

\bibitem{lai2011large}
Lai, K., Bo, L., Ren, X., Fox, D.: A large-scale hierarchical multi-view rgb-d
  object dataset. In: 2011 IEEE international conference on robotics and
  automation. pp. 1817--1824. IEEE (2011)

\bibitem{lee2018context}
Lee, D., Liu, S., Gu, J., Liu, M.Y., Yang, M.H., Kautz, J.: Context-aware
  synthesis and placement of object instances. In: Advances in Neural
  Information Processing Systems. pp. 10393--10403 (2018)

\bibitem{lin2014microsoft}
Lin, T.Y., Maire, M., Belongie, S., Hays, J., Perona, P., Ramanan, D.,
  Doll{\'a}r, P., Zitnick, C.L.: Microsoft coco: Common objects in context. In:
  European conference on computer vision. pp. 740--755. Springer (2014)

\bibitem{instancegithub}
{NVlabs}: Instance insertion,
  \url{https://github.com/NVlabs/Instance\_Insertion}

\bibitem{spadegithub}
{NVlabs}: Spade, \url{https://github.com/NVlabs/SPADE}

\bibitem{oquab2014learning}
Oquab, M., Bottou, L., Laptev, I., Sivic, J.: Learning and transferring
  mid-level image representations using convolutional neural networks. In:
  Proceedings of the IEEE conference on computer vision and pattern
  recognition. pp. 1717--1724 (2014)

\bibitem{park2019semantic}
Park, T., Liu, M.Y., Wang, T.C., Zhu, J.Y.: Semantic image synthesis with
  spatially-adaptive normalization. In: Proceedings of the IEEE Conference on
  Computer Vision and Pattern Recognition. pp. 2337--2346 (2019)

\bibitem{radford2015unsupervised}
Radford, A., Metz, L., Chintala, S.: Unsupervised representation learning with
  deep convolutional generative adversarial networks. arXiv preprint
  arXiv:1511.06434  (2015)

\bibitem{yolov3}
Redmon, J., Farhadi, A.: Yolov3: An incremental improvement. arXiv  (2018)

\bibitem{richter2022enhancing}
Richter, S.R., Al~Haija, H.A., Koltun, V.: Enhancing photorealism enhancement.
  IEEE Transactions on Pattern Analysis and Machine Intelligence  (2022)

\bibitem{richter2017playing}
Richter, S.R., Hayder, Z., Koltun, V.: Playing for benchmarks. In: Proceedings
  of the IEEE International Conference on Computer Vision. pp. 2213--2222
  (2017)

\bibitem{richter2016playing}
Richter, S.R., Vineet, V., Roth, S., Koltun, V.: Playing for data: Ground truth
  from computer games. In: European conference on computer vision. pp.
  102--118. Springer (2016)

\bibitem{ros2016synthia}
Ros, G., Sellart, L., Materzynska, J., Vazquez, D., Lopez, A.M.: The synthia
  dataset: A large collection of synthetic images for semantic segmentation of
  urban scenes. In: Proceedings of the IEEE conference on computer vision and
  pattern recognition. pp. 3234--3243 (2016)

\bibitem{rosario2018deep}
Rosario, G., Sonderman, T., Zhu, X.: Deep transfer learning for traffic sign
  recognition. In: 2018 IEEE International Conference on Information Reuse and
  Integration (IRI). pp. 178--185. IEEE (2018)

\bibitem{russakovsky2015imagenet}
Russakovsky, O., Deng, J., Su, H., Krause, J., Satheesh, S., Ma, S., Huang, Z.,
  Karpathy, A., Khosla, A., Bernstein, M., et~al.: Imagenet large scale visual
  recognition challenge. International journal of computer vision  115(3),
  211--252 (2015)

\bibitem{sadat2018effective}
Sadat~Saleh, F., Sadegh~Aliakbarian, M., Salzmann, M., Petersson, L., Alvarez,
  J.M.: Effective use of synthetic data for urban scene semantic segmentation.
  In: Proceedings of the European Conference on Computer Vision (ECCV). pp.
  84--100 (2018)

\bibitem{salimans2016improved}
Salimans, T., Goodfellow, I., Zaremba, W., Cheung, V., Radford, A., Chen, X.:
  Improved techniques for training gans. Advances in neural information
  processing systems  29 (2016)

\bibitem{sankaranarayanan2018learning}
Sankaranarayanan, S., Balaji, Y., Jain, A., Nam~Lim, S., Chellappa, R.:
  Learning from synthetic data: Addressing domain shift for semantic
  segmentation. In: Proceedings of the IEEE Conference on Computer Vision and
  Pattern Recognition. pp. 3752--3761 (2018)

\bibitem{schwarz2015rgb}
Schwarz, M., Schulz, H., Behnke, S.: Rgb-d object recognition and pose
  estimation based on pre-trained convolutional neural network features. In:
  2015 IEEE international conference on robotics and automation (ICRA). pp.
  1329--1335. IEEE (2015)

\bibitem{seib2020mixing}
Seib, V., Lange, B., Wirtz, S.: Mixing real and synthetic data to enhance
  neural network training--a review of current approaches. arXiv preprint
  arXiv:2007.08781  (2020)

\bibitem{shahroudy2016ntu}
Shahroudy, A., Liu, J., Ng, T.T., Wang, G.: Ntu rgb+ d: A large scale dataset
  for 3d human activity analysis. In: Proceedings of the IEEE conference on
  computer vision and pattern recognition. pp. 1010--1019 (2016)

\bibitem{shorten2019survey}
Shorten, C., Khoshgoftaar, T.M.: A survey on image data augmentation for deep
  learning. Journal of Big Data  6(1), ~60 (2019)

\bibitem{tremblay2018training}
Tremblay, J., Prakash, A., Acuna, D., Brophy, M., Jampani, V., Anil, C., To,
  T., Cameracci, E., Boochoon, S., Birchfield, S.: Training deep networks with
  synthetic data: Bridging the reality gap by domain randomization. In:
  Proceedings of the IEEE Conference on Computer Vision and Pattern Recognition
  Workshops. pp. 969--977 (2018)

\bibitem{unitygameengine}
{Unity Technologies}: Unity, \url{https://unity.com}

\bibitem{wang2018high}
Wang, T.C., Liu, M.Y., Zhu, J.Y., Tao, A., Kautz, J., Catanzaro, B.:
  High-resolution image synthesis and semantic manipulation with conditional
  gans. In: Proceedings of the IEEE conference on computer vision and pattern
  recognition. pp. 8798--8807 (2018)

\bibitem{wu2016learning}
Wu, J., Zhang, C., Xue, T., Freeman, B., Tenenbaum, J.: Learning a
  probabilistic latent space of object shapes via 3d generative-adversarial
  modeling. Advances in neural information processing systems  29 (2016)

\bibitem{yosinski2014transferable}
Yosinski, J., Clune, J., Bengio, Y., Lipson, H.: How transferable are features
  in deep neural networks? In: Advances in neural information processing
  systems. pp. 3320--3328 (2014)

\bibitem{bdd}
Yu, F., Chen, H., Wang, X., Xian, W., Chen, Y., Liu, F., Madhavan, V., Darrell,
  T.: Bdd100k: A diverse driving dataset for heterogeneous multitask learning
  (2018), \url{https://arxiv.org/abs/1805.04687}

\bibitem{zhu2017unpaired}
Zhu, J.Y., Park, T., Isola, P., Efros, A.A.: Unpaired image-to-image
  translation using cycle-consistent adversarial networks. In: Proceedings of
  the IEEE international conference on computer vision. pp. 2223--2232 (2017)

\end{thebibliography}

\end{document}